\documentclass[sigconf]{acmart}

\usepackage{booktabs} 
\usepackage{enumitem}
\usepackage{graphicx}
\usepackage{epsfig}
\usepackage{amssymb}
\usepackage{amsmath}
\usepackage{xcolor}
\usepackage{xspace}
\usepackage{bbm}
\usepackage{caption}
\usepackage{hyperref}
\usepackage{subcaption}
\usepackage{flushend}
\usepackage{soul}
\usepackage{todonotes}

\setlist[itemize]{leftmargin=*}

\makeatletter
\DeclareRobustCommand\onedot{\futurelet\@let@token\@onedot}
\def\@onedot{\ifx\@let@token.\else.\null\fi\xspace}

\def\eg{\emph{e.g}\onedot} 
\def\ie{\emph{i.e}\onedot} 
 
\def\etc{\emph{etc}\onedot} 
 
\def\etal{\emph{et al}\onedot}
\makeatother

\setcopyright{rightsretained}

\begin{document}
\title{Visual Search at eBay}

\author{Fan~Yang, Ajinkya Kale, Yury Bubnov, Leon Stein, Qiaosong Wang, Hadi Kiapour, Robinson~Piramuthu}
\affiliation{
\institution{eBay Inc. \\
\{fyang4, ajkale, ybubnov, lstein, qiaowang, mkiapour, rpiramuthu\}@ebay.com}
}

\renewcommand{\shortauthors}{F. Yang et. al.}

\begin{abstract}
In this paper, we propose a novel end-to-end approach for scalable visual search infrastructure. We discuss the challenges we faced for a massive volatile inventory like at eBay and present our solution to overcome those~\footnote{A demonstration video can be found at \url{https://youtu.be/iYtjs32vh4g}.}. 
We harness the availability of large image collection of eBay listings and state-of-the-art deep learning techniques to perform visual search at scale. Supervised approach for optimized search limited to top predicted categories and also for compact binary signature are key to scale up without compromising accuracy and precision. Both use a common deep neural network requiring only a single forward inference. The system architecture is presented with in-depth discussions of its basic components and optimizations for a trade-off between search relevance and latency. This solution is currently deployed in a distributed cloud infrastructure and fuels visual search in eBay ShopBot and Close5. We show benchmark on ImageNet dataset on which our approach is faster and more accurate than several unsupervised baselines. We share our learnings with the hope that visual search becomes a first class citizen for all large scale search engines rather than an afterthought. 
\end{abstract}

\copyrightyear{2017} 
\acmYear{2017} 
\setcopyright{acmcopyright}
\acmConference{KDD'17}{}{August 13--17, 2017, Halifax, NS, Canada.}
\acmPrice{15.00}
\acmDOI{http://dx.doi.org/10.1145/3097983.3098162}
\acmISBN{ISBN 978-1-4503-4887-4/17/08}

%
%




\maketitle

\section{Introduction}

\begin{figure}
\centering
\includegraphics[width=.9\linewidth]{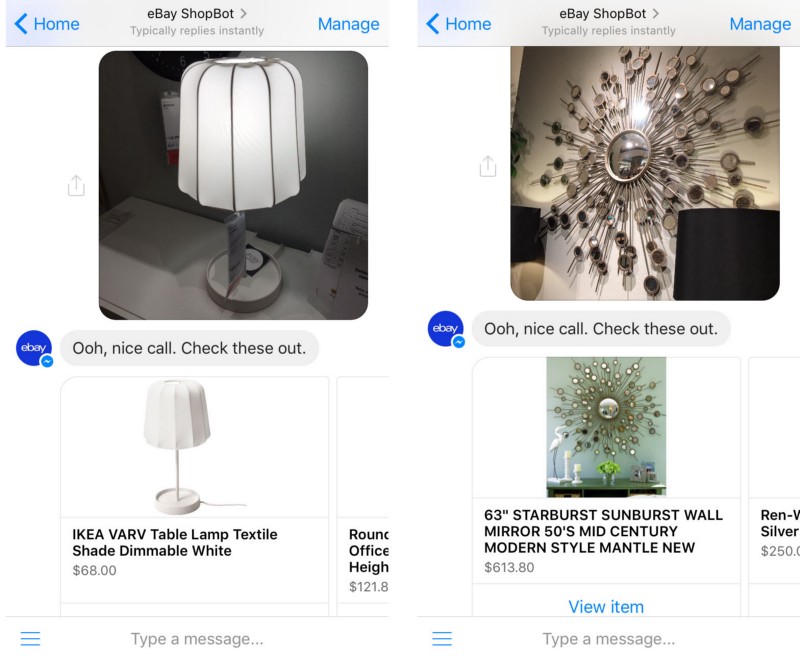}
\caption[]{Visual Search in eBay ShopBot\footnotemark. Accepts novel images as query. Note the quality of query images taken at store. Exact product was retrieved in these cases. Price comparison by taking a picture instead of bar code scanning or typing will become a common trend.} 
\label{fig:teaser}
\end{figure}
\footnotetext{\url{https://shopbot.ebay.com}}

With the exponential rise in online photos and openly available image datasets, visual search or content-based image retrieval (CIBR) has attracted a lot of interest lately. Although many successful commercial systems have been running visual search, there are very few publications describing the end-to-end system in detail, including algorithms, architecture, challenges and optimization of deploying it at scale in production~\cite{jing:pinterest,DBLP:journals/corr/KimKNKKJ16,DBLP:conf/sigir/WroblewskaR16}. 

Visual Search is an extremely challenging problem for a marketplace like eBay for 4 major reasons:
\begin{itemize}
\item \textit{Volatile Inventory}: Unlike the scenario for standard search engines, in a dynamic marketplace like eBay, numerous items are listed and sold every minute. Thus, listings are short-lived.
\item \textit{Scale}: Most solutions for visual search  work on small to mid scale datasets but fail to operate at eBay scale. In addition, eBay inventory covers numerous fine-grained categories that are difficult to classify. We need a distributed architecture to handle our massive inventory with high search relevance and low latency. 
\item \textit{Data Quality}: Image quality is diverse in eBay inventory since it is a platform that enables both high volume and occasional sellers. Compare this with catalog quality images in various popular commerce sites. Some listings may have incorrect or missing labels, which adds more challenges to model learning. 
\item \textit{Quality of Query Image}: eBay ShopBot allows users to upload query images taken from any source and not just limited to an existing image in the inventory (see Figure~\ref{fig:teaser}).
\end{itemize}

\begin{figure*}[ht]
\centering
\includegraphics[width=.9\linewidth]{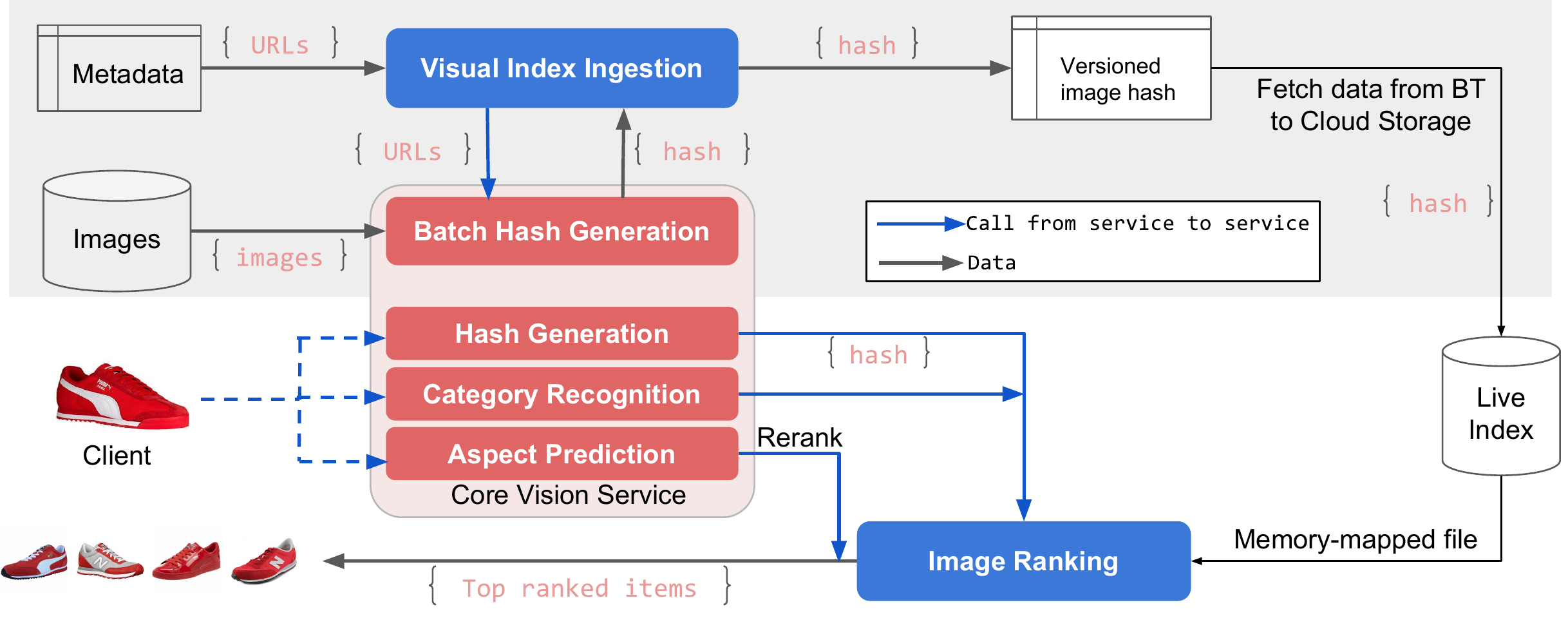}
\caption{Overview of our visual search system. The top part shows image ingestion and indexing. The bottom part illustrates the flow during inference. Our core service accepts a query image from the client, performs category/aspect prediction and hash generation in a single pass, and then sends the results to image ranking service (Section~\ref{sec:ranking}). The image ranking service takes the top predicted categories and hash of the query image to match against hash in live index to produce an initial recall set of retrieval results. These are further refined by aspect-based re-ranking for better semantic relevance (Section~\ref{sec:aspect}).}
\label{fig:overview}
\end{figure*}

We describe how we address the challenges above. We present our approach, tips and tricks for building and running a computationally efficient visual search system at scale. We will touch on the architecture of our system and how we leverage deep neural networks for precision and speed. Details of training a unified deep neural network (DNN) for category classification and binary signature extraction, along with aspect prediction are discussed.
We experiment with a public image dataset (ImageNet~\cite{DBLP:conf/cvpr/DengDSLL009}) to quantitatively evaluate the effectiveness of our network. We also showcase how visual search at eBay is deployed in a recently released product - eBay ShopBot, a chatbot integrated into Facebook Messenger platform\footnote{\url{https://shopbot.ebay.com/}}, and an eBay-owned location-based buying and selling mobile application -Close5\footnote{\url{https://www.close5.com}}, capable of discovering similar products from eBay inventory for users.

Rest of the paper is organized as follows: Section~\ref{sec:related} reviews recent literature on visual search algorithms based on deep learning and how our approach differs from existing commercial applications. 
This is followed by our approach of category/aspect prediction and binary hash extraction in Section~\ref{sec:algorithm}. We present our cloud-based distributed image search architecture in Section~\ref{sec:system}. Next, we show experimental results with quantitative analysis on ImageNet~\cite{DBLP:conf/cvpr/DengDSLL009} containing 1.2M images to prove the effectiveness of our model. Finally, results by eBay ShopBot and Close5 are presented in Section~\ref{sec:shopbot} and Section~\ref{sec:close5}, followed by conclusion in Section~\ref{sec:conclusion}.

\section{Related Work}
\label{sec:related}

Finding similar items using a seed image is a well-studied problem \cite{datta:imageretrieval}, but it remains a challenging one as images can be similar on multiple levels. Recently, with the rise of deep convolutional neural networks, visual search has attracted lot of interest \cite{jagadeesh:streetfashion,jing:pinterest,yamaguchi:parseclothing,bhardwaj:palettepower}. Deep learning has proven extremely powerful for semantic feature representation and multi-class image classification on large image datasets such as ImageNet \cite{krizhevsky:cnnimagenet}.

Krizhevsky \etal~\cite{krizhevsky:cnnimagenet} demonstrate strong visual retrieval results on ImageNet, where they use the last hidden layer to encode images and compute the distance between images in Euclidean space. This is infeasible for large scale visual search.
Various approaches utilize deep convolutional neural networks (CNNs) to learn a compact representation of images to achieve fast and accurate retrieval, while reducing the storage overhead. Most approaches~\cite{babenko:neuralcodes,DBLP:conf/aaai/XiaPLLY14,DBLP:conf/cvpr/LaiPLY15,DBLP:conf/aaai/ZhuL0C16,DBLP:conf/mm/GaoSZZS15,DBLP:conf/ijcai/LiWK16} focus on learning hash functions in a supervised manner using pairwise similarity of similar and dissimilar images so that the learned binary hashes capture similarity of images in the original Euclidean space. 
PCA and discriminative dimensionality reduction proposed by Babenko \etal~\cite{babenko:neuralcodes} provide short codes that give state-of-the-art accuracy, but require a large set of image pairs to minimize the distances between them.
Xia \etal~\cite{DBLP:conf/aaai/XiaPLLY14} propose a two-stage framework to learn hash functions as well as feature representations, but it requires expensive similarity matrix decomposition. 
Lai \etal~\cite{DBLP:conf/cvpr/LaiPLY15} directly learn hash functions with a triplet ranking loss while Zhu \etal~\cite{DBLP:conf/aaai/ZhuL0C16} incorporate a pairwise cross-entropy loss and a pairwise quantization loss, both under a deep CNN framework.
However, for a volatile inventory like eBay, where new products surface frequently, it is computationally inefficient and infeasible to collect a huge amount of image pairs across all categories, especially when the category tree is fine-grained and overlapping at times.

Alternatively, some works directly learn hash functions using point-wise labels without resorting to pairwise similarities of images~\cite{DBLP:conf/cvpr/LiongLWMZ15,lin:supbinaryhash}. Lin \etal~\cite{lin:supbinaryhash} suggest a supervised learning technique to avoid pairwise inputs and learn binary hash in a point-wise manner that makes it enticing for large-scale datasets. An unsupervised learning approach~\cite{DBLP:conf/cvpr/LinLCZ16} is also proposed to learn binary hashes by optimizing three types of objectives without utilizing image annotations. 
Despite of the success of these works, there are still challenges and unanswered questions about how to convert these research works into real production. Previous algorithms have been only evaluated on datasets containing at most few millions of images. However, given eBay scale datasets, it is challenging and non-trivial to perform hundreds of millions of hamming distance computations with low latency, while discovering the most relevant items at the same time.

With these realistic challenges in mind, we propose a hybrid scalable and resource efficient visual search engine. Our first contribution is a hybrid technique which uses supervised learning in three stages, first to classify an item image into the right category bucket to significantly cut down on the search space, followed by a supervised binary hashing technique proposed by Lin \etal~\cite{lin:supbinaryhash} and end with a re-ranking based on visual aspect (brand, color, \etc) recognition to ensure we show similar items not just based on the class or category but also across similar aspects. The first 2 tasks are performed using a single DNN at inference. The second contribution is to demonstrate the distributed architecture we deployed in eBay ShopBot to serve Visual Search in a highly available and scalable cloud-based solution. Some commercial visual search engines are heavily fashion-oriented~\cite{DBLP:journals/corr/KimKNKKJ16,DBLP:conf/sigir/WroblewskaR16}, where supported categories are limited to clothing products, while our goal is generic visual search and object discovery with significantly wider category coverage. We also learn binary image representations from deep CNN with full supervision instead of using low-level visual features~\cite{DBLP:conf/sigir/WroblewskaR16} or expensive floating point deep features~\cite{jing:pinterest}. In addition, we allow users to freely take photos in an unconstrained environment, which differentiates us from existing visual search systems that only accept clean, high-quality catalog images already in their inventory.





\begin{figure}[t]
\centering
\includegraphics[width=0.9\linewidth]{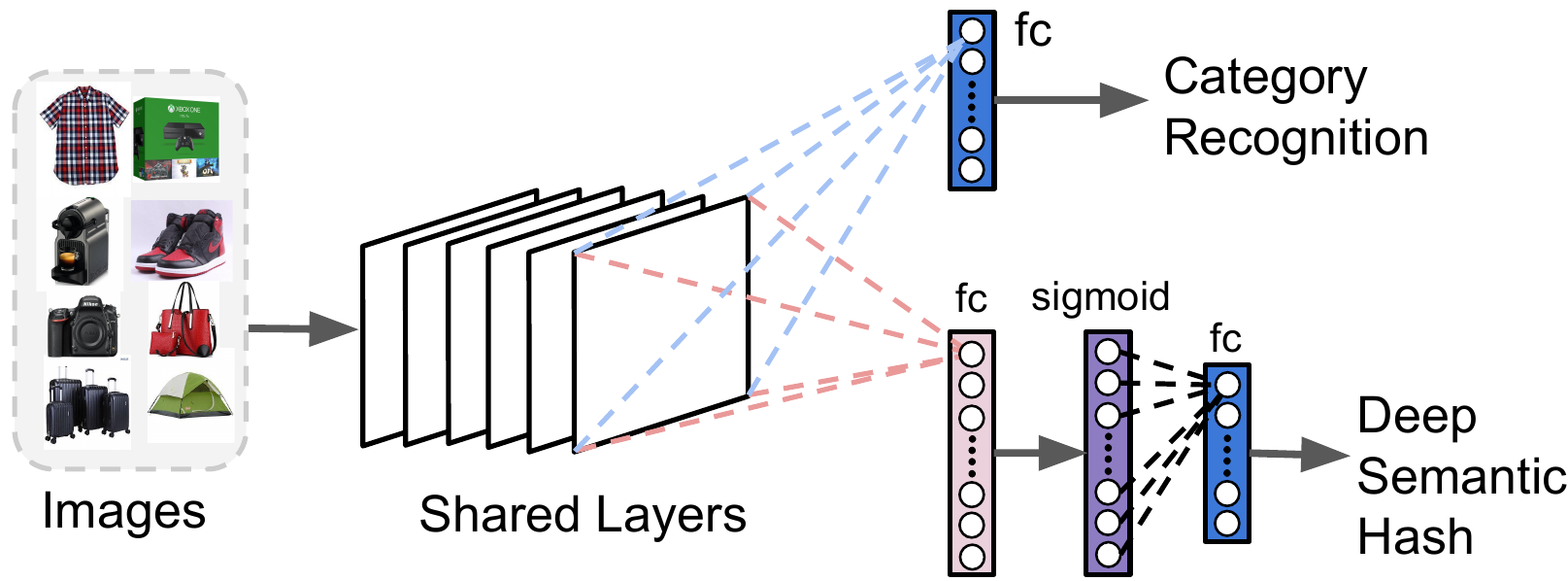}
\caption{Network Topology. We modified ResNet-50~\cite{DBLP:conf/cvpr/HeZRS16} with split topology for category prediction (top stream) and binary hash extraction (bottom stream). Sigmoid activations from the bottom stream are binarized during inference to get hash. Last layers in both streams were trained to predict category. This supervision helps with infusing semantic information in binary hash. Details in Section~\ref{sec:algorithm}.}
\label{fig:model}
\end{figure}

\section{Approach}
\label{sec:algorithm}
Figure~\ref{fig:overview} illustrates the overall system architecture of our visual search system. As in  \cite{jing:pinterest,DBLP:journals/corr/KimKNKKJ16}, our approach is based on a DNN. However, instead of directly extracting features from the DNN and performing exhaustive search over the entire database, we search only among top predicted categories and then use semantic binary hash with Hamming distance for fast ranking. For speed and low memory footprint, we use shared topology for both category prediction and binary hash extraction, where we use just a single pass for inference.

\subsection{Category Recognition}
\label{sec:leaf}

Rather than classifying objects into generic categories, such as persons, cats and clothes, \etc, we aim to recognize much finer object categories of  products. For example, we would like to distinguish men's athletic shoes from women's athletic shoes, maternity dresses from dancing dresses, or even coins from different countries. We refer to these fine-grained categories as leaf categories. We use state-of-the-art ResNet-50 network~\cite{DBLP:conf/cvpr/HeZRS16} for a good trade-off between accuracy and complexity. The network was trained from scratch based on a large set of images from diverse eBay product categories including fashion, toys, electronics, sporting goods, \etc. {We remove any duplicate images and split the dataset into a training set and a validation set, where every category has images in both training and validation sets. Each training image is resized to $256 \times 256$ pixels and the $227 \times 227$ center crop and its mirrored version are fed into the network as input. We used the standard multinomial logistic loss for classification tasks to train the network. To fully exploit the capacity of the deep network, we fine-tune the network with various combinations of learning parameters. Specifically, we change learning parameters after training for several epochs, and repeat this process several times until validation accuracy saturates.}

To further improve the network's ability to handle object variations, we also include on-the-fly data augmentation\footnote{\url{https://github.com/kevinlin311tw/caffe-augmentation}} during training to enrich training data, which includes random geometric transformations, image variations and lighting adjustments. Table~\ref{tab:augmentation} shows the absolute improvement by data augmentation against various image rotations. {All images used in this experiment are from the validation set, so they cover all the categories. All learning parameters are the same expect for the additional data augmentation module.}
Overall, data augmentation brings 2\% absolute improvement to top-1 accuracy in category prediction.
%

\begin{table}[t]
  \centering
  \caption{Robustness of data augmentation against image rotation. Numbers are absolute improvement of top-1 accuracy to predict category. Rotation angles are clockwise (in degree).}
  \small
  \vspace{-3mm}
  \renewcommand{\arraystretch}{1}
    \begin{tabular}{c|cccc|c}
    \hline
    Rotation angle & 0 & 90 & 180 & 270 & Mean \\
    \hline
    Improvement & 1.73\% &  2.04\% & 2.22\% & 1.98\% & 2.00\% \\
    \hline
    \end{tabular}%
  \vspace{-2mm}    
  \label{tab:augmentation}%
\end{table}
\subsection{Aspect Prediction}
\label{sec:aspect}

\begin{figure}[t]
\centering
\includegraphics[width=0.99\linewidth]{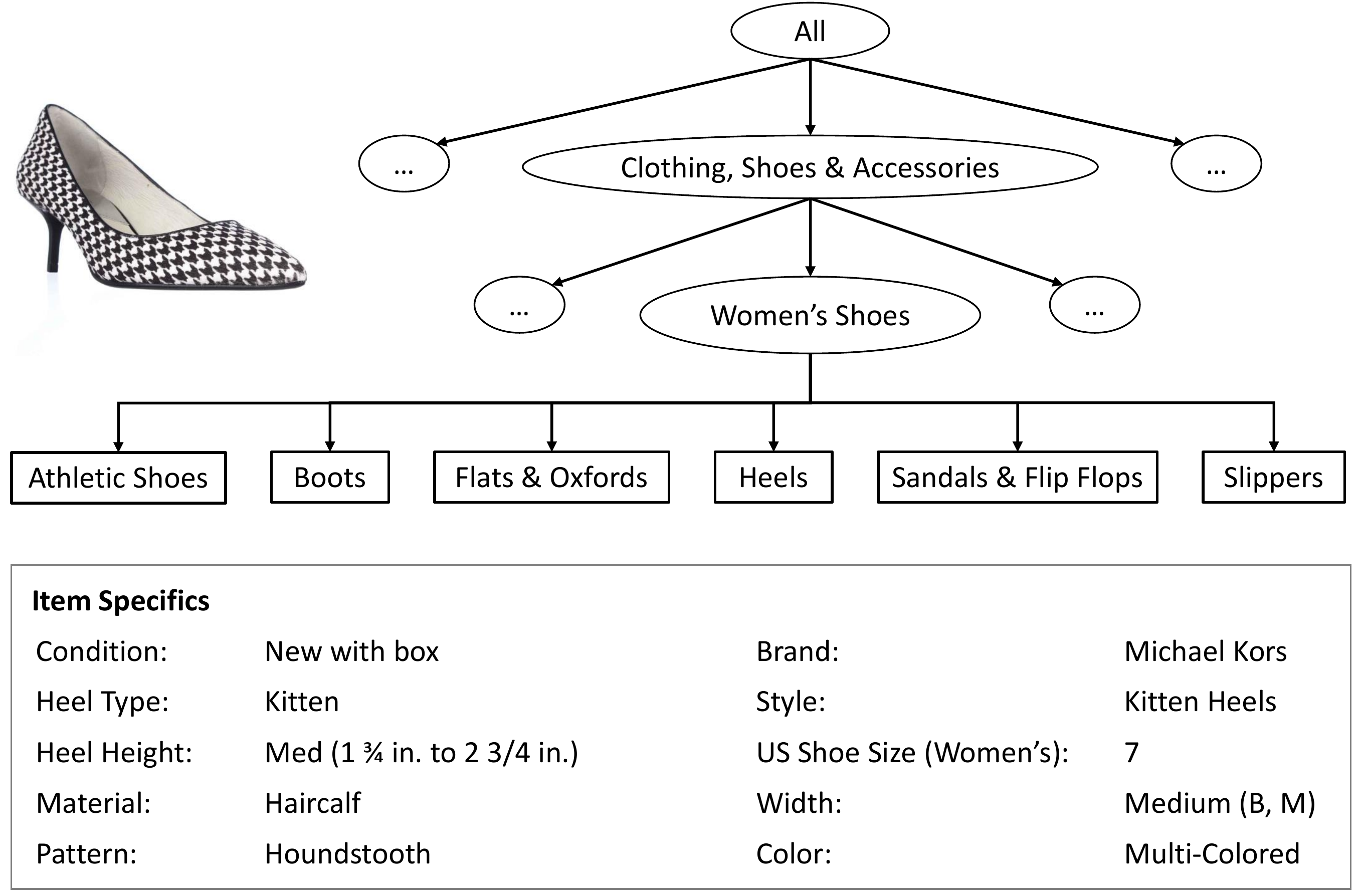}
\vspace{-2mm}
\caption{Simplified eBay Listing. Each listing has aspects (item specifics). The associated category is part of a category tree. We use leaf nodes for categories.}
\vspace{-4mm}
\label{fig:cat-tree}
\end{figure}

Product images are associated with rich information apart from the leaf category labels, such as color, brand, style, material, \etc. Such properties, called \emph{aspects} (Figure~\ref{fig:cat-tree}), add semantic information to each product. Our aspect classifiers are built on top of the shared category recognition network (Section \ref{sec:leaf}). Specifically, in order to save computation time and storage, all aspect models share the parameters with the main DNN model, up to the final \emph{pool} layer. Next, we create a separate branch for each aspect type. Our aspects cover a wide range of visual attributes including color, brand, style, \etc. Note that while some aspects such as color appear under multiple leaf categories, other aspects are specific to certain categories. \eg, precious stones are relevant to jewelry, while sleeve length is relevant to tops \& blouses. Therefore, we embed the image representation from \emph{pool5} layer with a one-hot encoded vector representation of leaf category. This integrates visual appearance and categorical information into one representation. This is particularly useful for our choice of multi-class classifiers. We use gradient boosted machines ~\cite{friedman:gbm} for their speed and flexibility in supporting both categorical and non-categorical input. Since the idea behind boosting is combining a series of weak learners, our model creates several splits of the visual space, according to the leaf categories. We use XGBoost~\cite{conf/sigkdd/ChenG} to train the aspect models. It allows for fast inference  with minimal resources using CPU only. We train a model for each aspect that can be inferred from an image. 

\subsection{Deep Semantic Binary Hash}
\label{sec:hashing}

Scalability is a key factor when designing a real world large-scale visual search system. Although we can directly extract and index features from some of the convolutional layers or fully-connected layer, it is far from optimal and does not scale well. It is both burdensome and costly to store real-valued feature vectors. In addition, computing pairwise Euclidean distance is inefficient, which significantly increases latency and deteriorates user experience. To address such challenges, we represent images as binary signatures instead of real values in order to greatly reduce storage requirement and computation overhead. We will show in Section \ref{sec:quantitative-category-prediction} that we do not lose much in accuracy in going from real valued to binary feature vector. 

Following~\cite{lin:supbinaryhash}, {we append an additional fully-connected layer to the last shared layer (Figure~\ref{fig:model}), and use sigmoid function as the activation function}. The sigmoid activation function limits the activations to bounded values between 0 and 1. {Therefore, it is a natural choice to binarize these activations by a simple threshold of $0.5$ during inference. Although more sophisticated binarization algorithms can be applied, we find that using 0.5 as the threshold already achieves balanced bits and leads to satisfactory performance in our production.}
This bottom stream of split topology in network is trained independently from the top stream, but by fixing the shared layer and learning weights for the later layers with the same classification loss that we use for category recognition. This supervised approach helps encode semantic information in the binary hash. We will show in Section~\ref{sec:quantitative-similarity-search} that our approach gives huge gains when compared to a popular unsupervised approach. Thus, we use a single DNN to predict category as well as to extract binary semantic hash. We also use this network to extract feature vector for aspect prediction. All of these operations are performed in a single pass. The complete model is presented in Figure~\ref{fig:model}.

We use 4096 bits in our binary semantic hash, {which corresponds to the number of neurons of the newly added fully-connected layer from the deep semantic hash branch in Figure~\ref{fig:model}}. Using 4096-dimensional binary feature vectors substantially reduces the storage requirement. Specifically, one such binary vector occupies only 512 bytes, making the total storage space under 100GB for 200M images. In contrast, {if extracting features from the last shared layer (\emph{pool5} layer), we obtain an 8192-dimensional floating point vector (from $2 \times 2$ feature maps of 2048 convolutional filters),} which requires 32KB space (using 32-bit floating point), resulting in an enormous storage of 6.1TB for the same amount of images. This gives over 90\% storage reduction. Moreover, it is far more efficient to use binary representation as Hamming distance is much faster than Euclidean distance to compute.

To further improve speed, we only search for the most similar images from the top predicted leaf categories, so that we can greatly reduce the overhead in an exhaustive search over the entire database covering all categories. Top matches returned from the top categories are merged and ordered again to generate the final ranked list.

\begin{figure}[t]
\centering
\includegraphics[width=.96\linewidth]{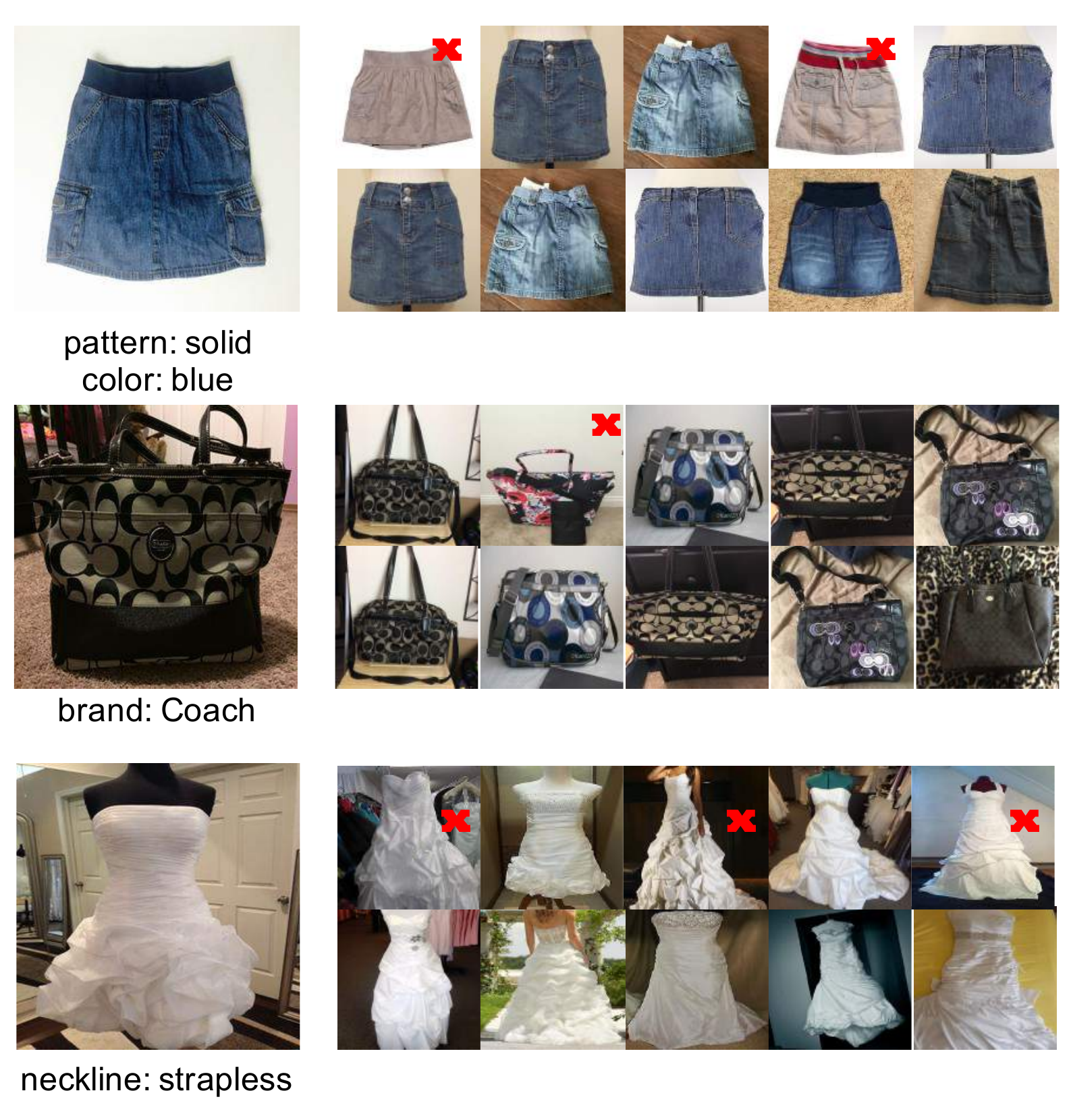}
\vspace{-3mm}
\caption{Refining search results by aspects. Predicted aspect values are shown. For each query, the $1^{st}$ and $2^{nd}$ rows present search results before \& after aspect-based re-ranking (Section~\ref{sec:aspect-based-reranking}), respectively. Red cross is shown on images with unmatched aspects as query.}
\vspace{-5mm}
\label{fig:aspect}
\end{figure}

\subsection{Aspect-based Image Re-ranking}
\label{sec:aspect-based-reranking}
Our initial results are obtained by only comparing binary signatures of images. However, we can further improve search relevance by utilizing semantic information from aspect prediction. Suppose our model generates $n$ aspects ($a_i^q$) for a query image $q$. Each matched item in the inventory has a set of $m$ aspects ($a_j$) and values as populated by the seller during listing. We check whether the predicted aspects match such ground-truth aspects and assign a ``reward point'' $w_i$ to each predicted aspect $a_i^q$ that has an exact match. The final score, defined as aspect matching score $S_{aspect} = \frac{1}{\sum_{i=1}^n w_i} \sum_{i=1}^n \sum_{j=1}^m w_i \mathcal{I}(a_i^q = a_j)$, is obtained by accumulating all scores of matched aspects.
Here $\mathcal{I}$ is an indicator function that equals 1 only when the predicted aspect and ground-truth aspect are the same. For simplicity, we assigned hard-coded reward points for all aspects, although they can be learned from data.  
In addition, rather than treating all aspects equally, we assign different reward points to them considering that some aspects are more important for users to make a purchase in e-commerce scenarios. In our system, we assign larger points to the aspects size, brand and price while equal importance for all other aspects.

After calculating the aspect matching score, we blend it with the visual appearance score $S_{appearance}$ (normalized Hamming distance) from image ranking to obtain the final visual search score to re-rank the initial ranked list of product images, \ie, effective similarity score $S = \lambda S_{appearance} + (1-\lambda) S_{aspect}$.
Linear combination allows fast computation without performance degradation. The combination weight $\lambda$ is fixed (0.75) in our current solution but is also configurable dynamically to adapt to changes over time. {We give more importance to appearance scores in order to be less sensitive to possible noise in aspect labels created by inexperienced sellers.}

Figure~\ref{fig:aspect} shows some examples where relevance is improved by aspect re-ranking. For the blue denim skirt, before re-ranking, the first and fourth retrieved images do not match the query in color. We observe that deep features, trained for category classification, might sometimes deprioritize color information. Color matches after re-ranking by aspect value for color. The second example contains a handbag in poor lighting conditions. Due to image quality, we could not get exact match of product. Instead, we get \emph{similar} bags. Without aspect prediction, the second image is not restricted to the brand in query. After re-ranking by aspects, brand in top retrieved images match with query. In the third example, straps of the wedding dress get overlooked in the hash representations. Re-ranking by aspects such as shape of the neckline refines the result to match fine-grained properties while preserving the overall similarity, even though we did not retrieve the exact product. 




\section{System architecture}
\label{sec:system}

\subsection{Image Ingestion and Indexing}
\begin{figure}[t]
\centering
\epsfig{file=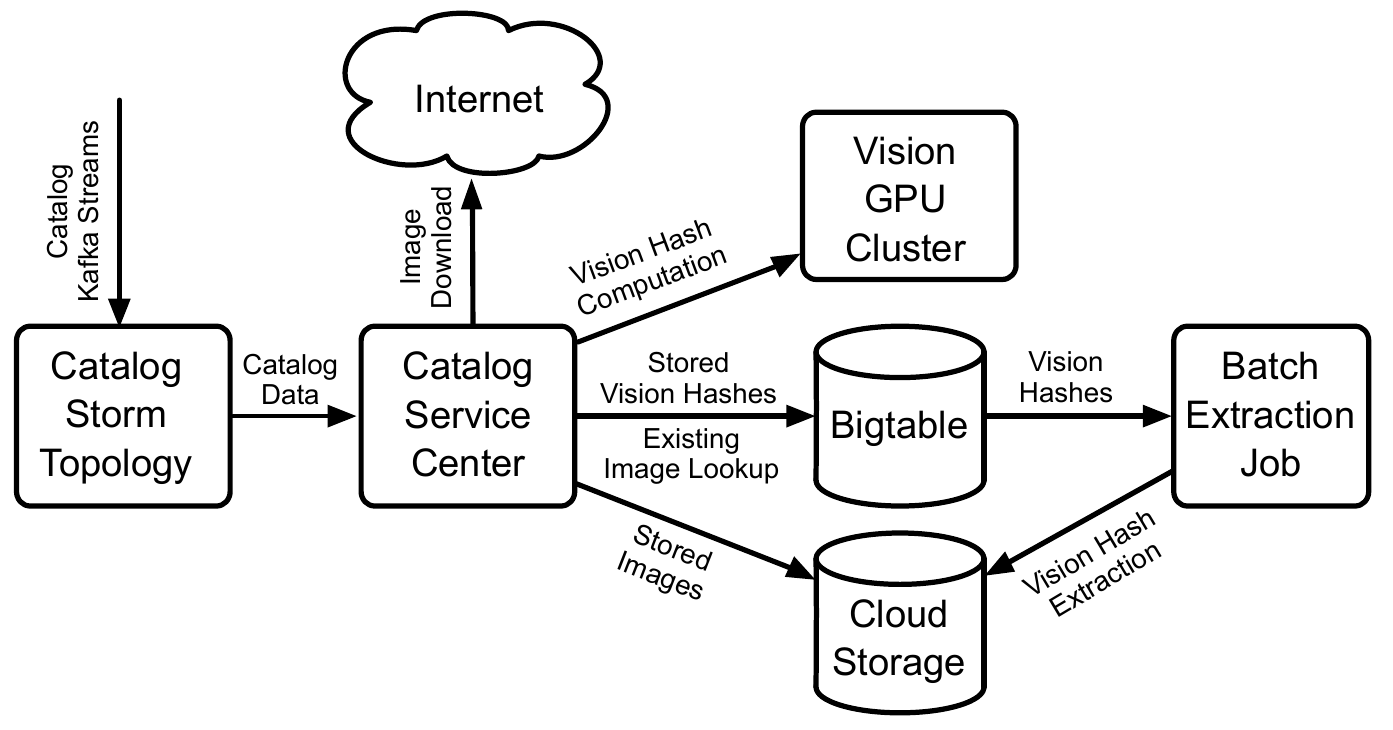, width=0.45\textwidth}
\vspace{-2mm}
\caption{Image ingestion system architecture}
\vspace{-2mm}
\label{fig:ingestion}
\end{figure}

eBay sellers worldwide generate numerous inventory image updates per second. Each listing may contain multiple images, and also multiple variations with multiple images. Our ingestion pipeline (Figure~\ref{fig:ingestion}) detects image updates in near-real-time and maintains them in cloud storage. To reduce storage requirements, duplicate images (about a third) across listings are detected and cross-linked. 

%
%
%
To detect duplicates, we compare MD5 hashes over image bits. Initially, we computed hash over bits of color channels after resizing and converting image to RGB model, however, it turned out that using exact matching, \ie, computing hash directly on image bits without resizing gives close duplicate detection rates, while being significantly cheaper computationally.

As images from new listings arrive, we compute image hashes for the main listing image in micro-batches against the batch hash extraction service (Figure~\ref{fig:overview}), which is a cluster of GPU servers running our pre-trained DNN models. Image hashes are stored in a distributed database (we use Google Bigtable), keyed by the image identifier.

For indexing, we generate daily image hash extracts from Bigtable for all available listings in the supported categories. The batch extraction process runs as a parallel Spark job in cloud Dataproc using HBase Bigtable API. The extraction process is driven by scanning a table with currently available listing identifiers along with their category IDs, and filtering listings in the supported categories. Filtered identifiers are then used to query listings from catalog table in micro-batches. For each returned listing, we extract the image identifier, and then lookup corresponding image hashes in micro-batches. The image hashes preceded by listing identifier are appended to a binary file. Both listing identifier and image hash are written with fixed length (8 bytes for listing identifier and 512 bytes for image hash). We write a separate file for each category for each job partition, and store these intermediate extracts in cloud storage. After all job partitions are complete, we download intermediate extracts for each category and concatenate them across all job partitions. Concatenated extracts are uploaded back to the cloud storage.

We update our DNN models frequently. To handle frequent updates, we have a separate parallel job that scans all active listings in batches, and recomputes image hashes from stored images. We keep up to 2 image hashes in Bigtable for each image corresponding to the older and the newer DNN model versions, so the older image hash version can be still used in extracts while hash re-computation is running.

\begin{figure}[t]
\centering
\epsfig{file=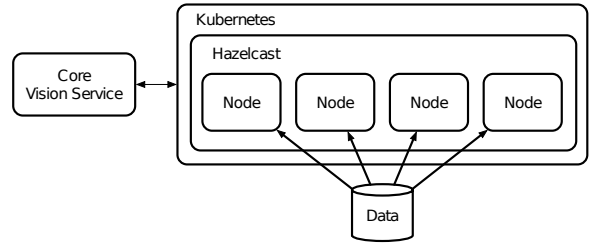, width=0.43\textwidth}
\vspace{-2mm}
\caption{Image ranking deployment}
\vspace{-4mm}
\label{fig:ranking}
\end{figure}

\subsection{Image Ranking}
\label{sec:ranking}
To build a robust and scalable solution for finding similar items across all items in the supported  categories, we create an image ranking service and deploy it in a Kubernetes\footnote{\url{https://en.wikipedia.org/wiki/Kubernetes}} cluster. Given the huge amount of data, we have to split image hashes for all the images across the cluster containing multiple nodes, rather than storing them on a single machine. This allows us to provide fault tolerance in case any single node becomes unavailable, and makes it possible to easily scale according to user traffic. In this scenario, each instance of the application will be responsible for a subset of the data, but collectively the cluster performs search across all the data. As our goal is to provide close-to-linear scalability, each node in the cluster should have knowledge about other nodes in order to decide which part of the data to serve. We use Hazelcast\footnote{\url{https://en.wikipedia.org/wiki/Hazelcast}} (an open source in-memory data grid) for cluster awareness.

When a node participates in the Hazelcast cluster, it receives notifications if other nodes are leaving or joining the cluster. Once the application starts, a ``cluster change'' event is received by every node. Then, each node checks current set of nodes. If there is a change, the data redistribution procedure is kicked off.  In this procedure, we split the data for each category into as many partitions as the number of nodes in the cluster. Each partition is assigned to a node in round robin fashion using a list of nodes sorted according to the ID assigned by Hazelcast. Starting node is determined by Ketama consistent hash\footnote{\url{https://en.wikipedia.org/wiki/Consistent_hashing}} from node ID. Thus, each node can generate the same distribution of data across the cluster and identify the part it is responsible for. To guarantee that all nodes have the same data, we leverage Kubernetes to share single disk, in read-only mode, across multiple pods. 

For initial discovery, during cluster startup, we leverage Kubernetes service with type ``Cluster IP''. When node performs DNS resolution of the service, it receives information about all nodes already in the Hazelcast cluster. Each node also periodically pulls information about other nodes from the DNS record to prevent cluster separation.

This design allows us to scale out to any number of nodes to satisfy different requirements, such as supporting increasing user traffic, decreasing search latencies by making each partition smaller and providing fault tolerance, \etc. If any single node becomes unavailable at exact moment when user sends request, search will be performed on live ones. Since data for each leaf category spreads across cluster, service availability can be guaranteed.

\begin{figure}
\centering
\epsfig{file=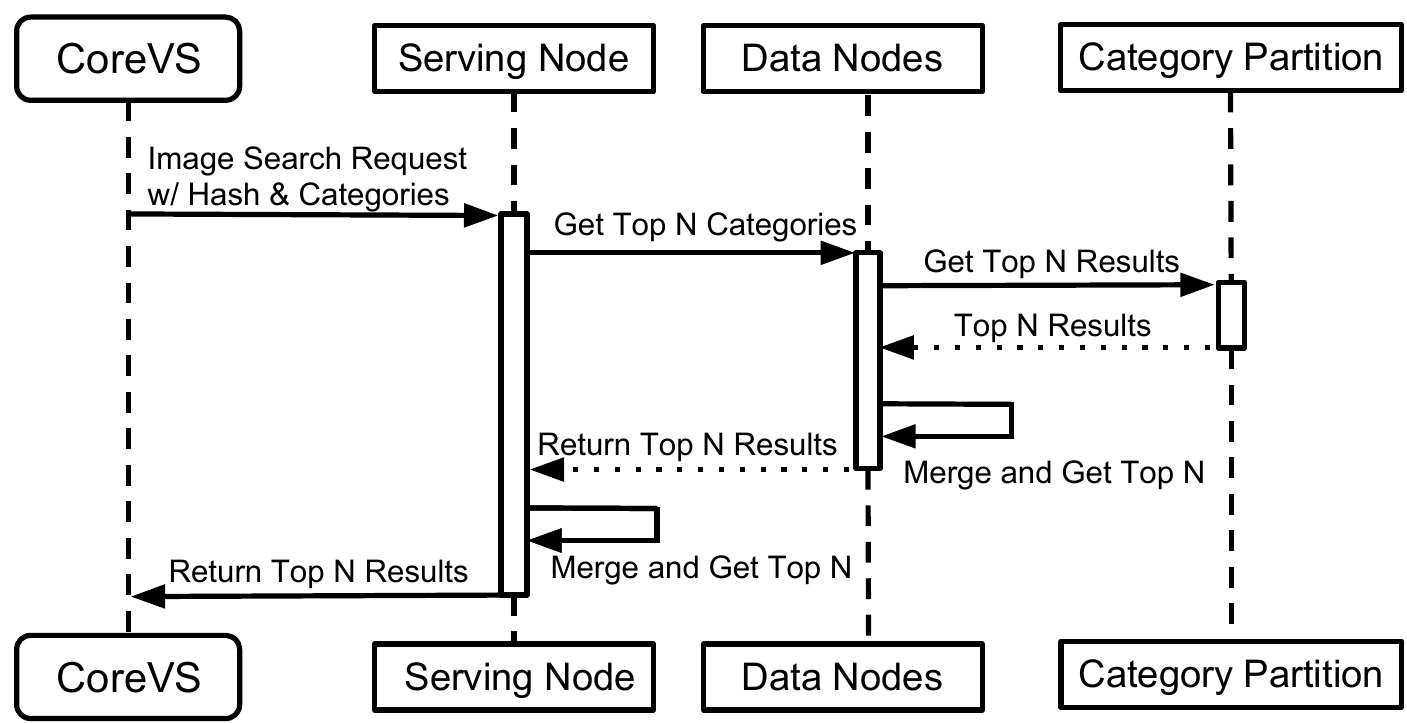, width=0.47\textwidth}
\vspace{-2mm}
\caption{Image ranking request sequence. CoreVS := Core Vision Service that predicts top categories, aspects and extracts binary semantic hash.}
\vspace{-4mm}
\label{fig:ranking-sequence}
\end{figure}

When our core vision service (Figure~\ref{fig:ranking}) sends incoming search requests to the image ranking service, the requests could be served be any node in the cluster, referred to as ``serving node''. Since each node is aware of the state of the cluster, it proxies incoming requests to all other nodes including itself. Each node further looks through partitions assigned to it and finds the closest $N$ listings for a given image hash if it has categories mentioned in the request (Figure~\ref{fig:ranking-sequence}). We use Hamming distance as the distance metric to discover the most similar listings. Data for each category is represented as a continuous array of listing IDs and corresponding image hashes. 
Each node divides category partition into a set of sub-partitions of the same number of available CPU cores and executes search in parallel to find the nearest $N$ items. 
Once the search is done in each sub-partition for each leaf category in the request, results are merged and the closest overall $N$ listings are returned. When search is completed on all data nodes, serving node performs similar merging procedure and sends back the result.

\section{Experiments on ImageNet}
\label{sec:imagenet}
\begin{figure}[t]
\centering
\includegraphics[width=0.9\linewidth]{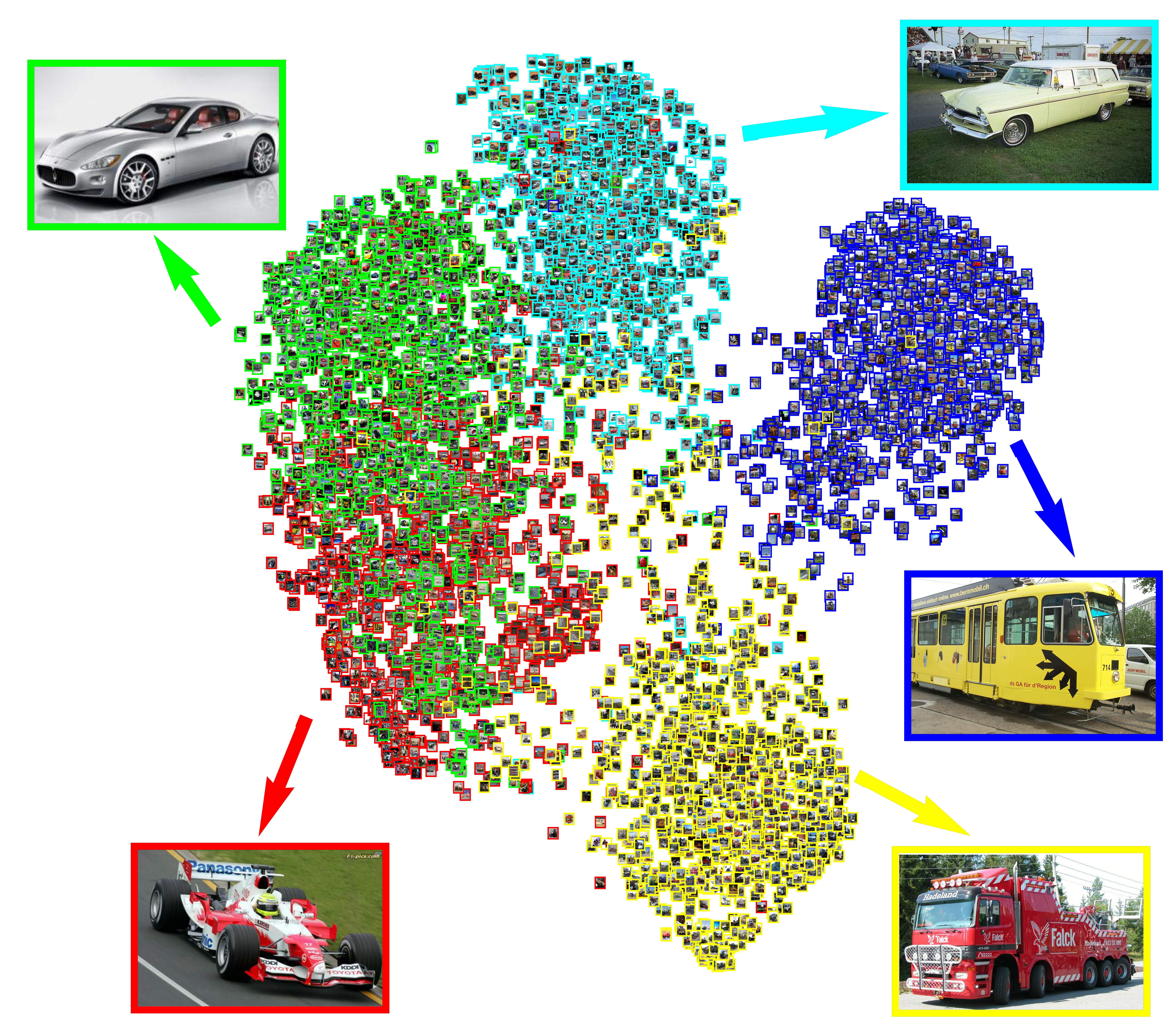}
\caption{t-SNE visualization~\cite{maaten2008visualizing} of 4096-bit binary semantic hash on ImageNet vehicle categories~(best viewed electronically and in color). This experiment is performed on 5 synsets, each containing 1300 images. We extract binary hashes for all images and apply t-SNE to obtain the image coordinates. Images belonging to the same category are marked by the same color. This illustrates that the binary hash preserves semantic information as well as similarity in the local neighborhood.}
\vspace{-4mm}
\label{fig:tsne}
\end{figure}

In this section, we conduct extensive experiments on the ImageNet~\cite{DBLP:conf/cvpr/DengDSLL009} dataset for reproducible proof of concept. We take the ResNet-50~\cite{DBLP:conf/cvpr/HeZRS16} model for category prediction and follow the same learning protocol in Section~\ref{sec:hashing} by fine-tuning a newly added hashing branch on ImageNet. Our classification branch (Figure~\ref{fig:model}) achieves top-1 validation error 25.7\% and top-5 validation error 7.9\%, which are only slightly higher than the accuracy of hashing branch: 24.7\% and 7.8\%, implying the discriminative power of the learned hash functions and capture of semantic content. We will confirm this quantitatively in the series of experiments to follow. Figure~\ref{fig:tsne} illustrates the embedding based on 4096-bit binary semantic hash for 5 synsets from ImageNet. This strengthens our claim qualitatively that the binary hash preserves semantic information and also local neighborhood. It is important to encode semantic information in hash to mitigate the undesirable effects of collision, since the items in collision will then be semantically similar.  

\begin{figure}[t]
\centering
\includegraphics[width=.94\linewidth]{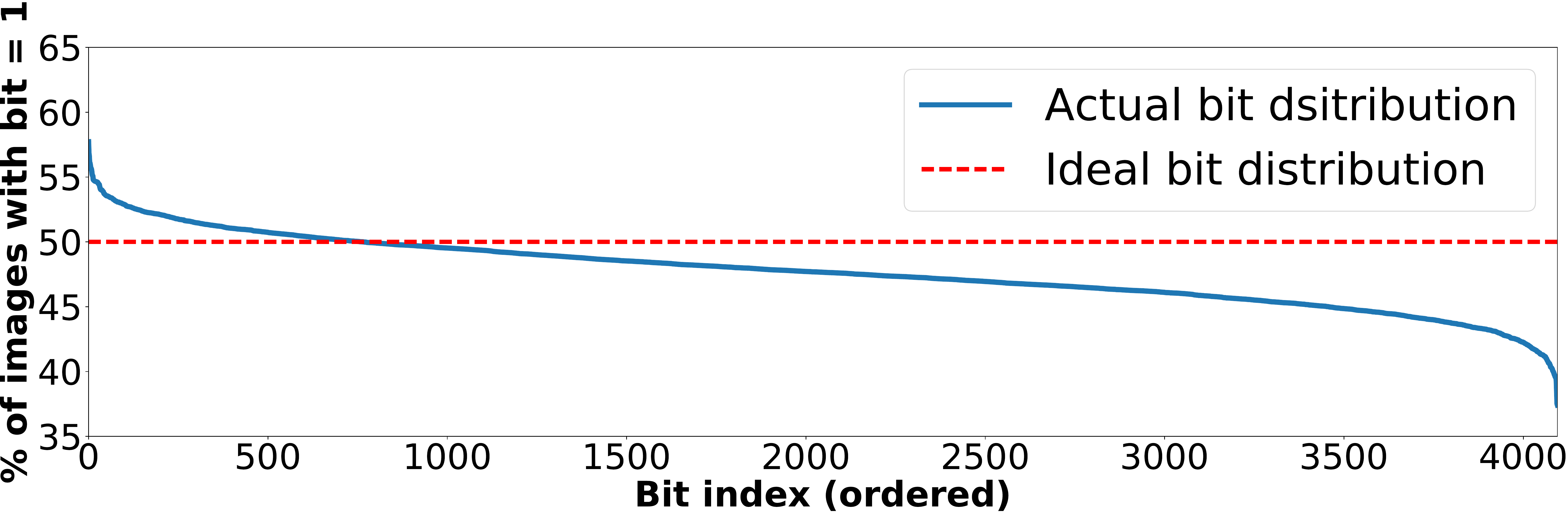}
\caption{Bit distribution on ImageNet~\cite{DBLP:conf/cvpr/DengDSLL009} training set. 84.1\% of bits are activate on 45\% to 55\% images. Ideal distribution for maximum entropy is uniform. Deviation from ideal is partly due to non-uniform diversity of data set and capture of semantic information. Details in Section~\ref{sec:bit-distribution}.}
\vspace{-2mm}
\label{fig:bit-distribution}
\end{figure}

\subsection{Bit Distribution}
\label{sec:bit-distribution}
One of the critical properties of learning good binary hash functions is to have balanced bits, \ie, forcing bits to be equally distributed over the full data. Given a specific bit of the binary hash, ideally it activates on half of the images in the training set while equals 0 on the remaining half. Such balanced distribution generates discriminative bits and reduces collision, as well as maintaining the largest entropy in the perspective of information theory, thus leading to more accurate matching for visual search.

In Figure~\ref{fig:bit-distribution}, we show the percentage of images from the training set where a bit is activated for each of the 4096 bits. While the curve is not perfectly flat at $50\%$ as in ideal case, it mostly lies around $50\%$, within the range from $45\%$ to $55\%$, which indicates that the bits learned from our model are roughly equally distributed. Specifically, as many as 3445 out of 4096 bits ($84.1\%$ of bits) activate on 45\% to 55\% images from the training set. This verifies the encoding efficiency of our binary semantic hash.

\begin{figure}[t]
\centering
\includegraphics[width=\linewidth]{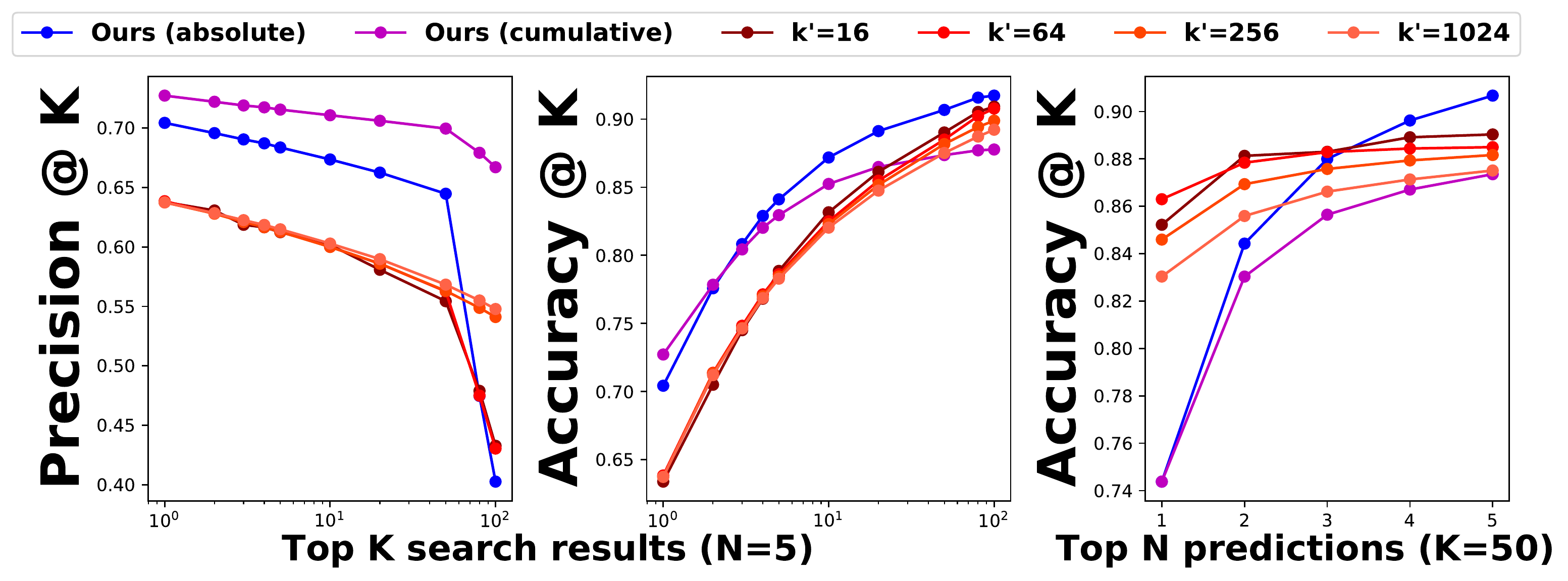}
\vspace{-2mm}
\caption{Category prediction based on top $K$ images. Performance of our approach and baseline by varying number of retrieved images $K$ and top $N$ category predictions. Here, $k'$ denotes number of centroids from k-means. Absolute top $N$ is better than top $N$ based on cumulative softmax confidence. Best accuracy for $N>3$. See Section~\ref{sec:quantitative-category-prediction} for details.}
\vspace{-4mm}
\label{fig:comp-cat}
\end{figure}

\subsection{Quantitative Comparison}

We conduct further experiments to evaluate the performance of our approach against a popular  unsupervised visual search, and show that our approach results in better retrieval results in terms of various evaluation metrics. We use the validation set of ImageNet~\cite{DBLP:conf/cvpr/DengDSLL009} as the query set to search for similar images from the training set. As a baseline, we perform k-means clustering ($k'=2^n$ where $n=4,6,8,10$) on binary hashes of images from the training set. Instead of first predicting $N$ class labels for each query image, we find the top $N$ nearest clustering centers first, and then search for $M$ most similar images within each cluster. In this way, we obtain $N \times M$ initial retrieved images, from which $K$ images are returned as the final search results.

\subsubsection{Category prediction}
\label{sec:quantitative-category-prediction}
Using predicted categories during search, our supervised approach produces more accurate results. We use precision@$K$ and accuracy@$K$ by varying the number of top $K$ returned results to evaluate the performance of baselines and our approach. These metrics measure the number of relevant results, \ie, from the same class as the query, and the classification accuracy, \ie, the query is considered to be correctly classified if there is at least one returned image belonging to the same class among the top $K$ retrieved results. For all experiments, we set $M=50$ so that we only search for 50 similar images within each predicted category. We have 2 flavors for our approach. First is to look at absolute top $N$ categories. The second is to use predictions up to a maximum of $N$ until we get a cumulative softmax confidence of 0.95. Our results show that the former is better. In Figure~\ref{fig:comp-cat}, accuracy of our approach rapidly improves as $K$ increases when $K \le 100$ while precision only drops slightly, which clearly shows that our approach does not introduce many irrelevant images into the top search results. Compared to different variants of the baseline, we are able to achieve better performance when retrieving as few as only 20 images. On the other hand, when we search within more predicted categories, the accuracy significantly improves as we include more candidates and outperforms all baseline variants. 

\subsubsection{Similarity search} 
\label{sec:quantitative-similarity-search}
\begin{figure}[t]
\centering
\includegraphics[width=0.45\textwidth]{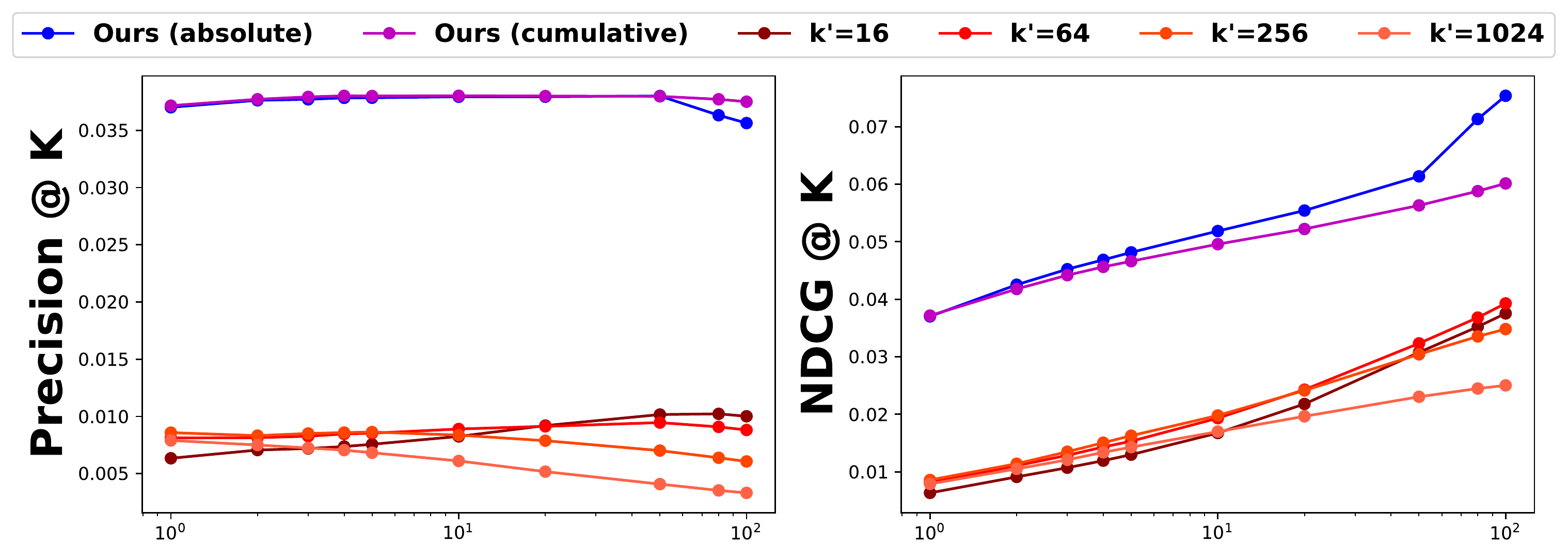}
\includegraphics[width=0.45\textwidth]{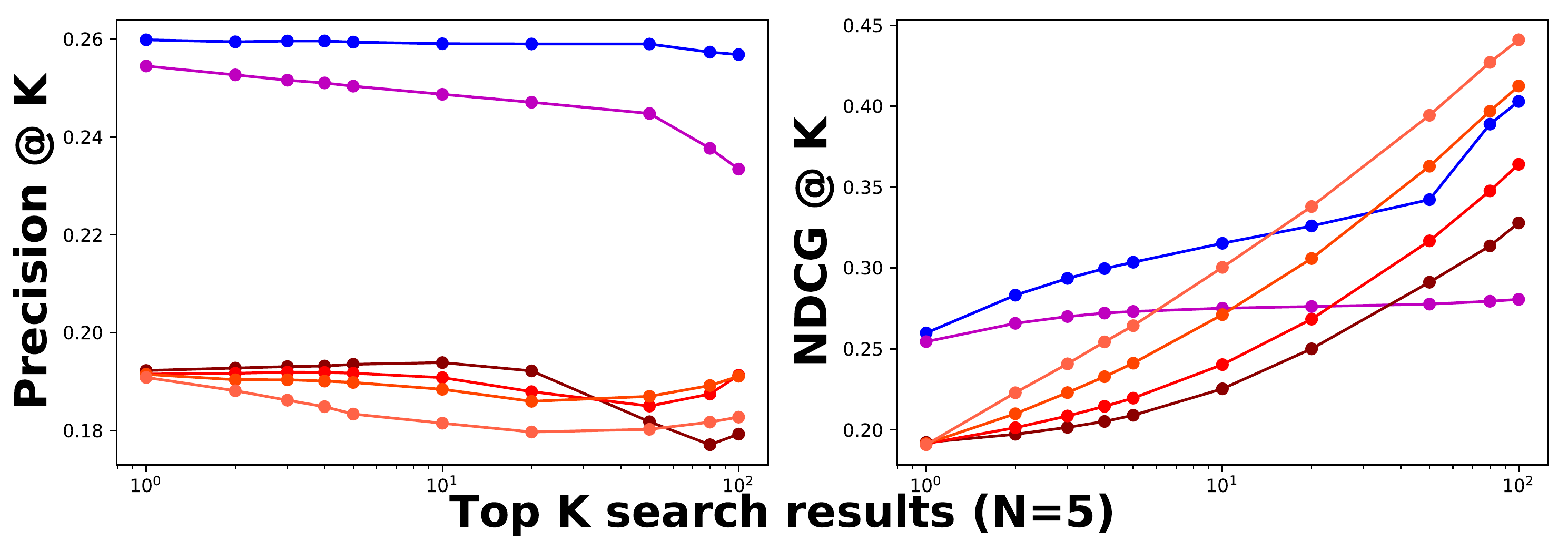}
%
\vspace{-2mm}
\caption{Search relevance. The top and bottom rows use a different ground-truth for nearest neighbor (NN). The former uses NNs in the entire data set and the latter is restricted to NNs from ground truth category of query. Note high relevance from our approach even though search is limited to only the top predicted categories. Absolute top $N$ is again better than top $N$ based on cumulative softmax confidence. See Section~\ref{sec:quantitative-similarity-search} for details.}
\vspace{-4mm}
\label{fig:comp-similarity}
\end{figure}

We further compare the performance of our approach and baseline in terms of similarity search, where we consider the results of exact $K$ nearest neighbors of a query as ground-truth and evaluate how the compared methods approximate the ground-truth. Therefore, we use normalized discounted cumulative gain (NDCG) to measure the quality (relevance) of the ranked list considering the ranks, and again precision@$K$. 
By taking into account the category information of images, we have another ground-truth to evaluate these methods by looking at the relevance at category-level. Our approach again outperforms baseline variants under both of the two scenarios (see Figure~\ref{fig:comp-similarity}).
\begin{table}[t]
  \centering
  \caption{Comparison of search ranking time per query (in ms), averaged over all queries, by our approach and baselines with various number $k'$ of k-means centroids.}
  \small
  \vspace{-3mm}
  \renewcommand{\arraystretch}{1}
    \begin{tabular}{c|ccccc}
    \hline
    Method & Ours & $k'=16$ & $k'=64$ & $k'=256$ & $k'=1024$ \\
    \hline
    $N=1$ & 33.5 &  2073.4 & 605.9 & 124.7 & 47.1\\
    $N=5$ & 150.6 & 8293.0 & 2246.4 & 720.9 & 172.3\\
    $N=10$ & 323.7 &  16252.1 & 4190.0 & 1056.1 & 325.3 \\
    \hline
    \end{tabular}%
\vspace{-5mm}    
  \label{tab:timing}%
\end{table}

\subsubsection{Timing} 

We also compare speed (excluding network forward pass) of our approach and several variations of the baseline. We run experiments with different $N$, \ie, top predictions or nearest k-means centroids and present the results in Table~\ref{tab:timing}. Our method is much more efficient than the baseline since we reduce the search space significantly by only searching the top $N$ predicted categories. Specifically, our method is $1.14\times$ faster than the baseline with a large $k'=1024$ when $N=5$, while being more accurate. Note that all experiments are conducted in Python with single-thread processing on a desktop, while the production environment (for live eBay inventory) has been extensively optimized (Section~\ref{sec:ranking}) to greatly reduce latency (Section~\ref{sec:latency}).

\section{Application: eBay ShopBot}
\label{sec:shopbot}

Our visual search was recently deployed in eBay ShopBot. Since then, visual search has become an indispensable part of  users' shopping journey and has served numerous customers. {People can find the best deals from eBay's 1 billion-plus live listings by uploading a photo. Early user data from ShopBot has shown that the number of shopping missions that started with a photo search has doubled since the launch of eBay ShopBot beta}.
In the following, we summarize two main scenarios where visual search is used in ShopBot.

\subsection{User query} 
eBay ShopBot allows users to freely take a photo (from camera or photo album) and find similar products in eBay's massive inventory. This is invaluable specifically when it is hard to precisely describe a product solely in words. {We do not set any restrictions on how the photos are taken, and can} support a wide range of queries from professional quality images to low quality user photos. Figure~\ref{fig:teaser} shows examples of user interface on ShopBot. Our visual search successfully finds exactly matching products despite noisy background and low quality query images.

\subsection{Anchor search}
\label{sec:anchor}
\begin{figure}[t]
\centering
\includegraphics[width=0.4\linewidth]{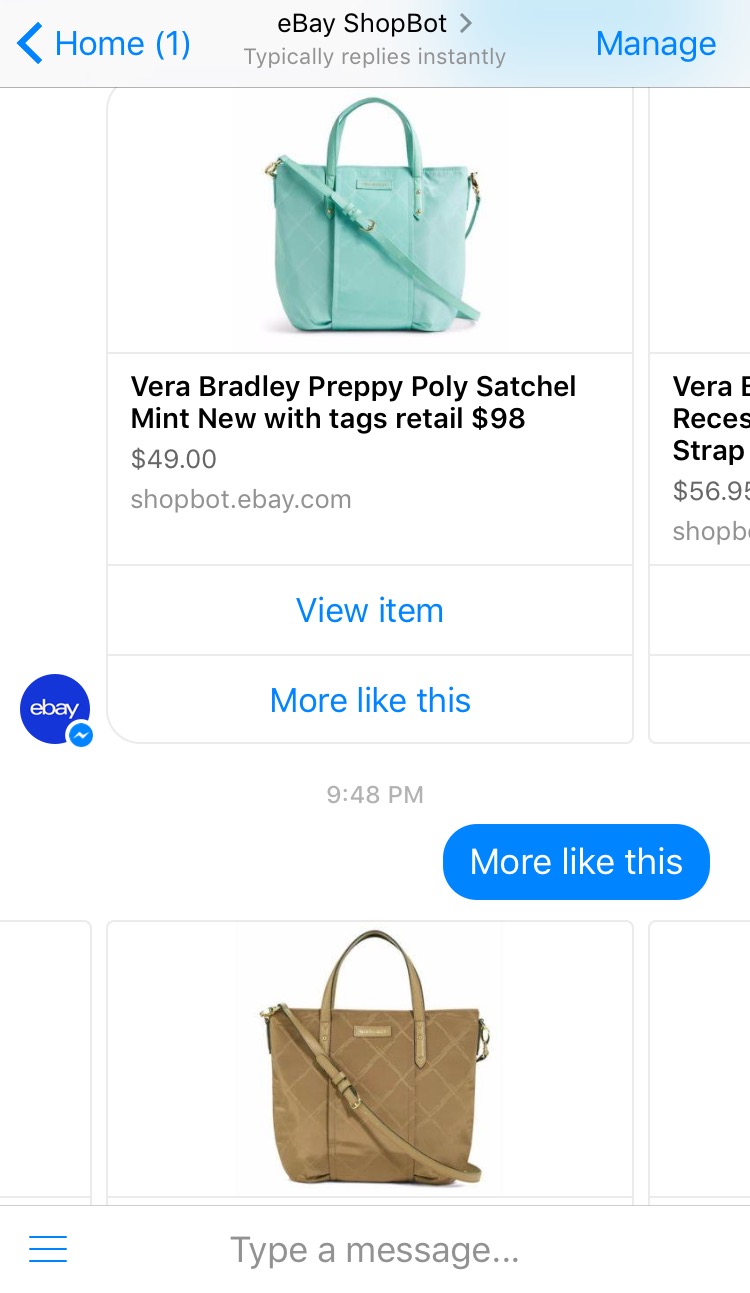}\hfill
\includegraphics[width=0.4\linewidth]{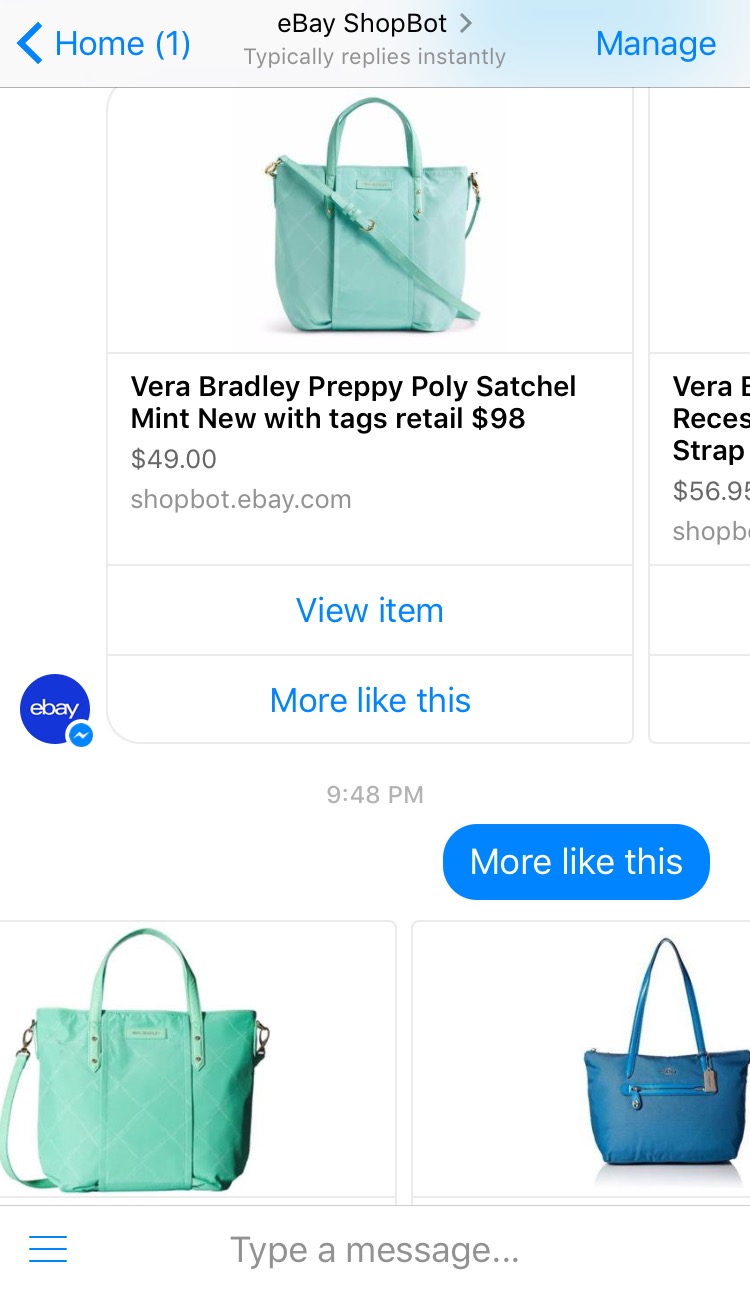}
\vspace{-3mm}
\caption{Anchor search. Query is from active eBay listing (not user-uploaded). Images on the right were shown when scrolled. See Section~\ref{sec:anchor} for details.}
\vspace{-5mm}
\label{fig:more-like-this}
\end{figure}

The second use case of visual search in eBay ShopBot is anchor search. Every time a user is presented with a list of products (even if she used text to initiate the search), she can click on the ``more like this'' button on any of the returned products to refine or broaden initial searches, and to find visually similar items. Here, the photos from the selected item serve as the anchor to initialize a new visual search. Since the anchor is actively listed, we take advantage of its meta information, such as category and aspects, to guide visual search even further. {In this way, we provide users with a non-linear search and browsing experience, which allows more flexibility and freedom. This feature has been a very popular feature with users with engagement increasing by 65\% since it became available.}
Figure~\ref{fig:more-like-this} shows an example where the user is looking for inspirations by finding visually similar handbags.

\subsection{Qualitative results}
\begin{figure*}[t]
\centering
\includegraphics[width=.75\linewidth]{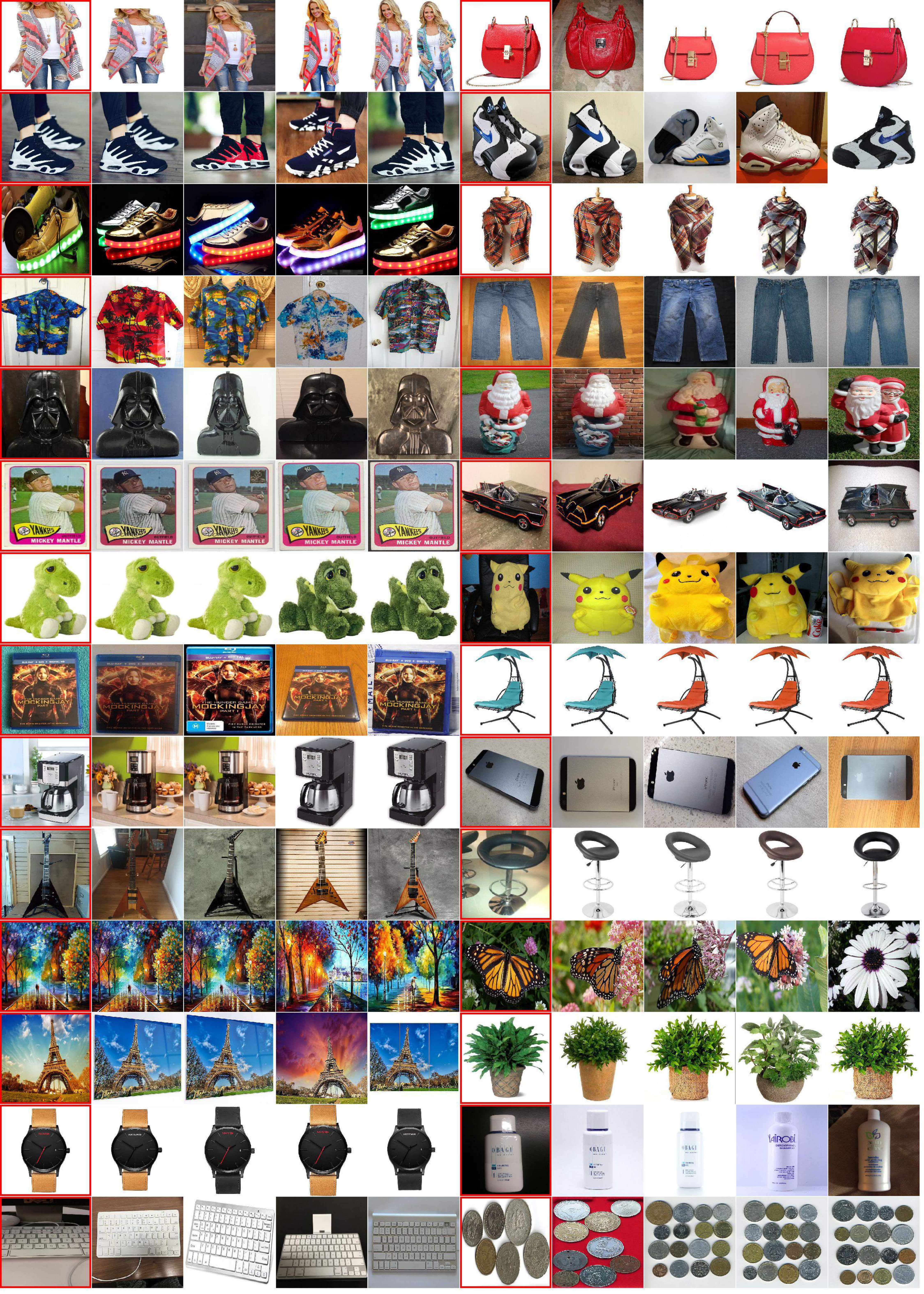}
\vspace{-2mm}
\caption{Qualitative results of our visual search. Query images (not taken from eBay site) have red border and are followed by top 4 ranked images from active eBay listings. Note the diversity in categories and image quality. When exact product is not found, retrieved results still share common semantics with query.}
\vspace{-4mm}
\label{fig:qualitative}
\end{figure*}

Since we could not share eBay dataset due to proprietary information, we present only qualitative results in Figure~\ref{fig:qualitative}. {Our visual search engine successfully discovers visually similar images from the massive and dynamic inventory of eBay despite the variety of categories, diverse composition and illumination. Note that all retrieved images are from active listings of eBay at the time of writing}. 

\subsection{Latency}
\label{sec:latency}
We report the latency of several main components. By extensive optimization and leveraging the computational power of cloud, the batch hash generation service takes 34ms per image on a single GPU (Tesla K80). In ShopBot scenario, given a user query, the deep network takes 125ms on average to predict the category, recognize aspects and generate image hash. The image ranking service only takes $25$ms and $70$ms to return 50 and 1000 items, respectively, depending on the size of each category. The aspect re-ranking only takes 10ms to re-rank as many as 1000 results. Therefore, the total latency is only a couple of hundreds milliseconds, plus miscellaneous overhead, which provides users with a fast and enjoyable shopping experience. 

\section{Application: Close5}
\label{sec:close5}

{Close5 is an eBay-owned platform for buying and selling locally. Users can freely take photos of and add descriptions to their products they want to sell to create a listing on Close5, which can be viewed by nearby users. Our visual search solution has been deployed and integrated with Close5 native application support millions of Close5 users. It has been extensively used in the following two scenarios.}

\subsection{Auto categorization}
\label{sec:category-close5}
{While most listings have user-created descriptions, some only have photos taken by users as title and description are not mandatory. Therefore, these listings without descriptions are not search-able by text, resulting in $200\times$ less chance of selling than others. In this case, our visual search predicts the category of the product based on images to populate category-related textual information, so that these listings can be search-able. Early statistics show that the average view of a listing increased by 21\% compared to those without descriptions and no auto categorization. 
}

\subsection{Similar items on eBay}
\label{sec:similar-close5}
\begin{figure}[t]
\centering
\includegraphics[width=0.48\linewidth]{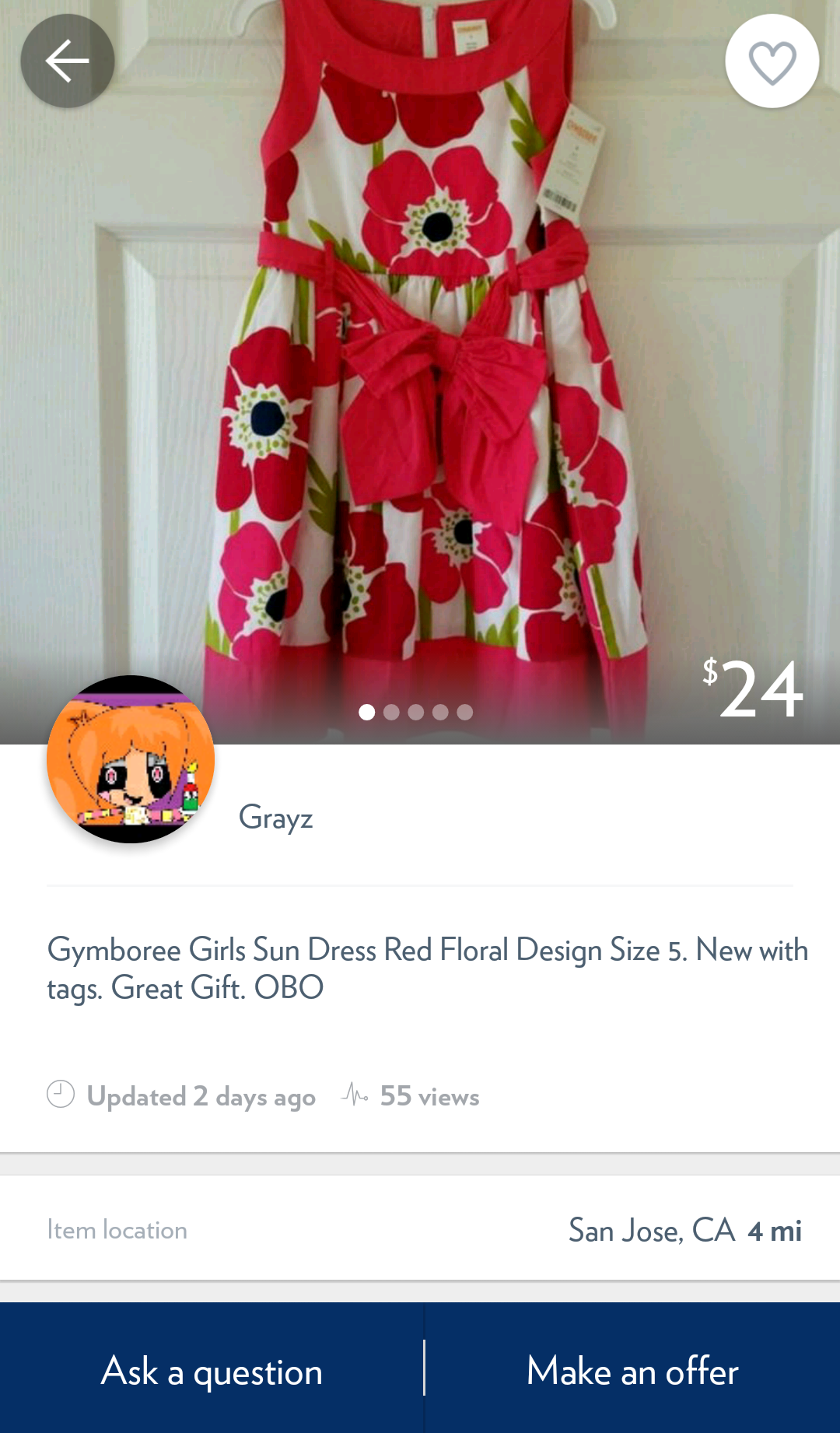}\hfil
\includegraphics[width=0.48\linewidth]{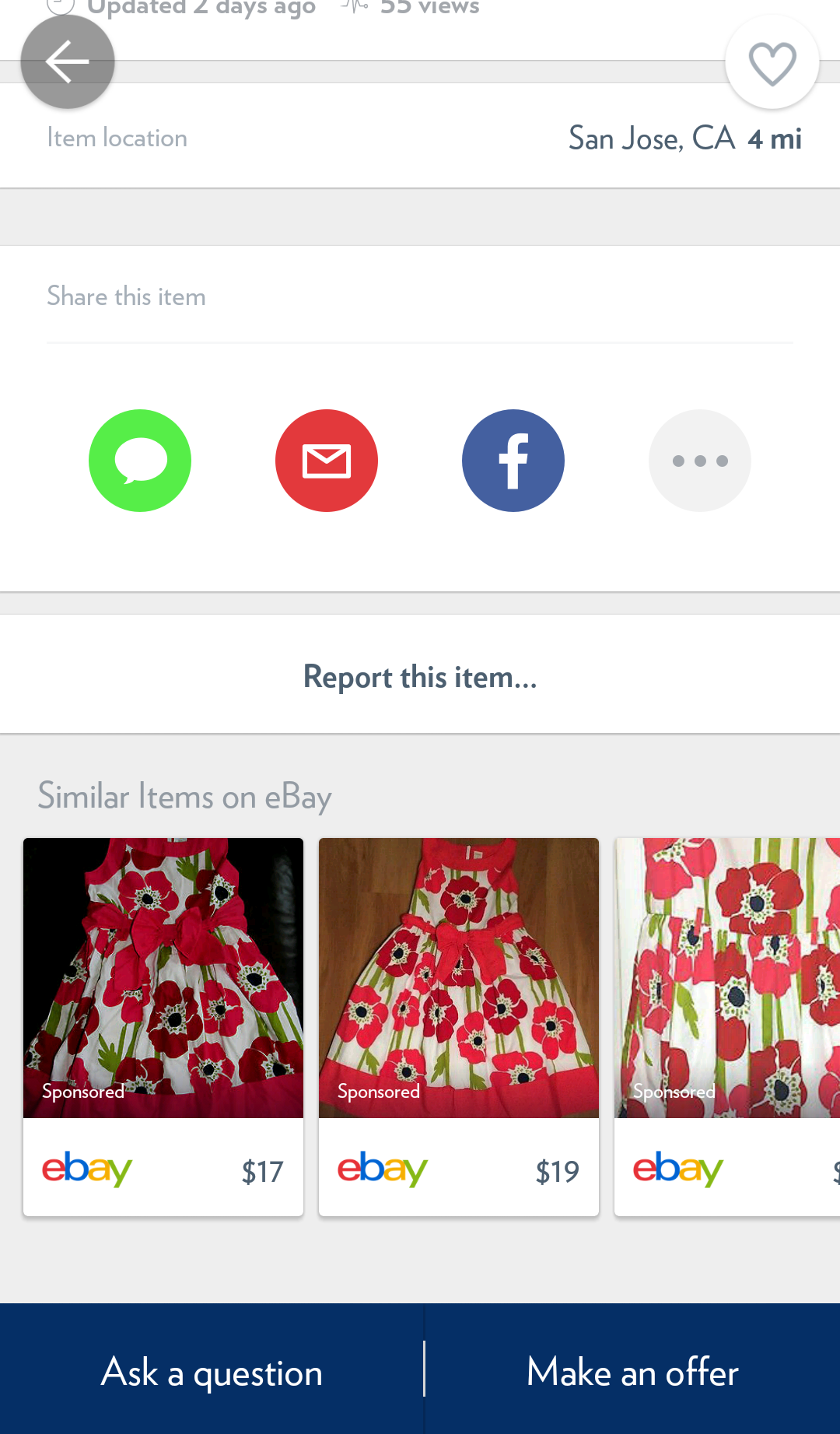}
\vspace{-3mm}
\caption{Screenshot of a listing from Close5. ``Similar items on eBay'' are provided by our visual search. Note that we find an exact match listed by a different seller at ebay.com, which has different inventory than close5.com.}
\vspace{-3mm}
\label{fig:close5}
\end{figure}

{To enable product discovery from eBay inventory for Close5 users, we integrate the visual feature functionality as a feature called ``Similar items on eBay''. When a user browses a listing, visual search automatically triggers with the first image from this Close5 listing as query. Results are presented at the bottom of the listing page with links directing users to eBay website. As shown in Figure~\ref{fig:close5}, our visual search successfully finds the same floral dress, as well as other similar dresses. With this feature, eBay inventory is exposed to millions of Close5 users, which provides both Close5 users and eBay sellers with improved buying and selling experience. Click through rate doubled since we rolled out this user using visual search.
}

\section{Conclusion}
\label{sec:conclusion}
We presented a scalable visual search infrastructure that leverages the power of deep neural network and cloud-based platform for efficient product discovery given a massive and volatile inventory like eBay. Highlights include searching only among images from top predicted categories, single DNN with split topology to predict category as well as extract compact and efficient binary semantic hash, aspect-based re-ranking to reinforce semantic similarity. We have also presented the system architecture and discussed several optimizations for a trade-off between search relevance and latency. Extensive experiments on a large public dataset verifies the discriminative ability of our learned model. {Additionally, we show that our visual search solution has been deployed successfully to the recently launched eBay ShopBot, and integrated into eBay-owned Close5 native application.}


%

\vspace{-4mm}
\bibliographystyle{ACM-Reference-Format}
\bibliography{vissearch} 

\end{document}